\documentclass[twoside]{article}
\usepackage[accepted]{aistats2016}

 \usepackage{relsize}
 \usepackage{enumitem}
\usepackage{wrapfig}
\usepackage{balance}
\usepackage{array}
\usepackage{amsmath}
\usepackage{mathrsfs}
\usepackage{amssymb}
\usepackage{amsthm}
\usepackage{dsfont}
\usepackage{tikz}
\usepackage{color}
\DeclareMathOperator*{\argmax}{arg\,max}

\newtheorem{theorem}{Theorem}
\newtheorem{definition}{Definition}
\newtheorem{corollary}{Corollary}
\newtheorem{remark}{Remark}

\newcommand{\EE}[1]{\mathbb{E}\left[#1\right]}

\newcommand*{\MyDef}{\mathrm{\tiny def}}
\newcommand*{\eqdefU}{\ensuremath{\mathop{\overset{\MyDef}{=}}}}
\newcommand*{\eqdef}{\mathop{\overset{\MyDef}{\resizebox{\widthof{\eqdefU}}{\heightof{=}}{=}}}}

\newcommand{\cG}{\mathcal{G}}

\newcommand{\bM}{{\bf M}}

\newcommand{\nothere}[1]{}

\usepackage{xspace}

\newcommand{\UCB}{\texttt{UCB}\xspace}
\newcommand{\MOSS}{\texttt{MOSS}\xspace}

\newcommand{\UCBN}{\texttt{UCB-N}\xspace}
\newcommand{\GraphMOSS}{\texttt{\textcolor[rgb]{0.5,0.2,0}{GraphMOSS}}\xspace}
\newcommand{\UCBmaxN}{\texttt{UCB-MaxN}\xspace}

\newcommand{\BARE}{\texttt{\textcolor[rgb]{0.5,0.2,0}{BARE}}\xspace}
\newcommand{\ELP}{\texttt{ELP}\xspace}

\newcommand{\expix}{\texttt{Exp3-IX}\xspace}
\newcommand{\expset}{\texttt{Exp3-SET}\xspace}
\newcommand{\expdom}{\texttt{Exp3-DOM}\xspace}

\usepackage{graphicx} 
\usepackage{natbib}
\usepackage{algorithm}
\usepackage{algorithmic}

\definecolor{babyblue}{rgb}{0.54, 0.81, 0.94}
\definecolor{citrine}{rgb}{0.89, 0.82, 0.04}
\usepackage[colorinlistoftodos, textwidth=22mm, shadow, textsize=small]{todonotes}

\newcommand{\rkdual}{r_k^{\circ}}
\newcommand{\rktdual}{r_{k,t}^{\circ}}
\newcommand{\rkprimetdual}{r_{k',t}^{\circ}}
\newcommand{\rkttdual}{r_{k,t+1}^{\circ}}
\newcommand{\rkprimettdual}{r_{k',t+1}^{\circ}}
\newcommand{\ridual}{r_i^{\circ}}
\newcommand{\rstardual}{r_\star^{\circ}}

\newcommand{\Ddualset}{\mathcal D^{\circ}}

\usepackage[colorlinks,
           linkcolor=red,
           citecolor=blue,
           urlcolor=magenta,
           linktocpage,
           plainpages=false]{hyperref}

\begin{document}

%

%

\twocolumn[

\aistatstitle{Revealing graph bandits for maximizing local influence}

\aistatsauthor{Alexandra Carpentier \And Michal Valko}

\aistatsaddress{Universit\" at Potsdam \And  SequeL team, Inria Lille - Nord Europe} ]

\begin{abstract}
We study a graph bandit setting where the objective of the learner is to detect the \emph{most influential node} of a graph by requesting as little information from the graph as possible. 
One of the relevant applications for this setting is marketing in social networks, where the marketer aims at finding and taking advantage of the most influential customers.
The existing approaches for bandit problems on graphs require either partial or complete knowledge of the graph. In this paper, we do not assume any knowledge of the graph, but we consider a setting where it can be gradually discovered in a \emph{sequential and active way.} At each round, the learner chooses a node of the graph and the only information it receives is a \emph{stochastic} set of the nodes that the chosen node is currently influencing. 
To address this setting, we propose \BARE, a bandit strategy for which we prove a regret guarantee that scales with the \emph{detectable dimension}, a problem dependent quantity that is often much smaller than the number of nodes.
\end{abstract}

\section{Introduction}

\emph{Bandit problems on graphs}~\citep{mannor2011from,caron2012leveraging} are sequential decision problems with \emph{limited feedback}, where the learner 
can take advantage of a \emph{graph structure} of the actions. This allows the learner to attain faster learning rates when compared to treating all the nodes independently. The recent popularity of this setting is due to its applications in marketing and advertising. In a typical case, the graph represents a social network, where the nodes are users and the edges encode the intensity of the social links between them. One marketing application that we target is  \emph{product placement}. An advertiser can offer a product to some  users in a hope that they will recommend the product to their contacts, i.e.,~to the neighboring nodes in the social network. The advertiser then observes the set of contacts that these users have influenced and that have  bought the product. The objective of the advertiser is to target \textit{influential users}, the nodes of the graph whose influence is the most important. Ideally, the advertiser would only offer products to the users with maximal influence.

What plays the key role, when it comes to effective detection of the most influential users? It is the graph structure, as it gives \textit{side information}, in particular, on the proximity between the nodes and their \textit{influence} on others. The learner can leverage this side information and \emph{learn faster}. The magnitude of the gains in the learning \emph{rates} naturally depends on the graph. Consequently, the performance guarantees can be expressed with graph quantities such as the \emph{clique partition number}~\citep{caron2012leveraging}, 
the \emph{independence  number} \citep{alon2013from}, or 
the \emph{minimum dominating set}~\citep{buccapatnam2014stochastic}. 
Furthermore, there are many models of influence and some of the known ones were introduced in the seminal work on spreading 
the influence through a social network~\citep{kempe2003maximizing,kempe2015maximizing}.
In the present paper, we consider \emph{local influence}, where a node on the graph influences
only its \emph{immediate neighborhood}.

Most of the existing approaches for active learning on graphs assume that either the \textit{entire graph}  is known in advance, or at least that a substantial \emph{part of the graph} is revealed to the learner after it selected the node. Typically, the algorithms require at least the knowledge of the set of neighbors of the neighbors of the nodes (\emph{second neighborhood}). 
This knowledge of the graph is crucial for existing learning algorithms \citep{mannor2011from,yu2011unimodal,caron2012leveraging,cesa-bianchi2013gang, alon2013from,gentile2014online,kocak2014efficient,gu2014online,valko2014spectral,
alon2015online,wu2015online,kocak2016online} to help them learn faster than in the case if no structure existed.
However, in some realistic scenarios, the graph information is \emph{not available}
to the learner beforehand. Typically, the operator of the social network would not freely reveal the social links and therefore the graph is not known to the advertiser. On the other hand, for instance in the advertising example presented above, the advertiser has some local access to the social network in the sense that it can get information of the set of users that were influenced to purchase products through the other targeted customers. This information can be gathered through \emph{promotional codes} when the goal is the \emph{product purchase}  or through ``likes''  in the case of  an \emph{information campaign}~\citep{caron2012leveraging}.

However, the existing graph bandit approaches do not allow to treat this scarce side information setting. 
Therefore, with the known tools, one can either (i) first thoroughly explore the graph and then apply existing graph bandit strategies, or (ii)  forget about the underlying graph structure and apply existing multi-arm bandit algorithms to the nodes of the graph. In both cases, it is necessary that the learner substantially explores the graph and therefore \emph{samples many nodes}, if not all of them. This is not very reasonable, for instance, in our marketing example, since graphs corresponding to social networks are usually  large. Moreover, the advertiser is unlikely to have a large enough budget to target all the nodes of the graph in order to learn which ones are the most  influential.


More formally, in this paper we consider a sequential graph learning problem where \emph{at 
the beginning, there is no information} about the underlying graph. The edges are only \emph{revealed progressively} as a result of the choices of the learner. Specifically, the learner observes the set of nodes that have been influenced by the chosen node. The objective of the learner is to find and target the \emph{most influential node} of the graph. 
We consider a \emph{local influence structure}. In this simple model, each node can only influence its neighbors. In this paper, we aim at finding a strategy for this problem that would  be practical on very large graphs and in particular, that does not scale with the number of nodes if the graph has some structure.
We propose a learning strategy that we call \BARE, a bandit revelation algorithm. The performance guarantees for this algorithm do not scale with the number of nodes, but with the \textit{detectable dimension}, a quantity that is often much smaller than the number of nodes. Specifically, \BARE does not require to sample the entire graph when the \textit{detectable dimension} is small. 

\emph{Stochastic bandits on graphs} were inspired by~\cite{mannor2011from} and were first studied 
by~\cite{caron2012leveraging} that proposed \UCBN and \UCBmaxN that closely follow \UCB, but in addition, they use side observations for better reward estimates (\UCBN) or choose one of the neighboring nodes with a better empirical estimate (\UCBmaxN). These improvements enable to improve guarantees, i.e.,~the regret does not scale with the  
number of nodes but  with the \emph{clique partition number}. 
Later, \cite{buccapatnam2014stochastic} improved the results of~\cite{caron2012leveraging} with LP-based solutions and  
guarantees scaling only with the \emph{minimum dominating set}.
Spectral bandits~\citep{valko2014spectral, kocak2014spectral,gu2014online} assume that each node
of the known graph has a mean reward that is \emph{smooth} on the graph, 
which means that the connected nodes give similar rewards. 
Moreover, the gang of bandits~\citep{cesa-bianchi2013gang} 
addresses the problem of multiple users, where each of the nodes 
possesses a contextual linear bandit itself and the linear weight vectors
of neighboring nodes are assumed to be similar.
Furthermore, online clustering of bandits~\citep{gentile2014online} assumes that the nodes
of the graph can be clustered with respect to some unknown underlying clustering
and the nodes within a cluster exhibit similar behavior. 
Yet another assumption of a special graph reward structure is exploited by unimodal bandits~\citep{yu2011unimodal,combes2014unimodal}
and networked bandits~\citep{fang2014networked}.

Bandits with side observations were first studied
in the more difficult \emph{non-stochastic setting}~\citep{mannor2011from}, where the rewards do not follow a fixed distribution. Their first algorithm, \ELP, comes with the guarantee expressed as a function of the \emph{clique number}. \ELP  was later followed by \expset~\citep{alon2013from}, \expdom~\citep{alon2013from}, and \expix~\citep{kocak2014efficient}, whose guarantees were proved to be functions of the \emph{independence number}, which gives either equal or a better guarantee. This line of work was recently extended to the setting beyond bandit feedback~\citep{alon2015online}, where the learner may not  observe the reward of the chosen node
and to \emph{noisy} side observations \citep{wu2015online,kocak2016online}.

The common feature of all prior approaches is the need of having access to knowledge of the portions of the graph, in order to get faster learning rates. These portions are larger than what our setting permits.
In particular, in our setting, the information revealed is just the \emph{first neighborhood} of the chosen node.

Recently, \cite{lei2015online,chen2015combinatorial,vaswani2015influence} investigated the combinations of offline influence maximization 
approaches with multi-arm bandit strategies for the online influence maximization 
in the independent cascade model of \cite{kempe2003maximizing} with \emph{semi-bandit} feedback.
%
Finally, another set of approaches considered restricted exploration constraints on a graph, 
modeling the crawling of the network
\citep{bnaya2013bandit,bnaya2013social,singla2015information}.


\section{Local influence bandit settings}
\label{sec:setting}
\subsection{Description of the problem}\label{ss:prob}
Let $\cG$ be a graph with $d$ nodes. When a node $i$ is selected, it can influence the nodes of $\cG$, including itself.  Node $i$ influences each node $j$ with \emph{fixed} but \emph{unknown} probability $p_{i,j}$. Let $\bM = (p_{i,j})_{i,j}$ be the $d\times d$ matrix that represents $\cG$. 

We consider the following online, active setting. At each round (time) $t$, the learner chooses a node $k_t$ and observes which nodes are influenced by $k_t$, i.e., the set $S_{k_t,t}$ of influenced nodes is \emph{revealed}. Let us also write $S_{k_t,t}(r)$ for the $r$th coordinate of $S_{k_t,t}$, i.e.,\@ it is~$1$ if~$k_t$ influences $r$ at time $t$ and $0$ otherwise. 
Given a budget of  $n$ rounds, the objective is to maximize the number of \emph{influences} that the selected node exerts. Formally, our goal is to find the strategy maximizing the performance
\[L_n = \sum_{t = 1}^n \left|S_{k_t,t}\right|.\]
The \textit{influence} of node $k$, i.e., the expected number of nodes that node $k$ exerts influence on, is by definition
\[r_k = \EE{\left|S_{k,t}\right|} = \sum_{j \leq d} p_{k,j}.\]
We also define the \textit{dual influence} of node $k$ as
\[\rkdual = \sum_{j \leq d} p_{j,k}.\]
This quantity is the expected number of nodes that exert influence on node $k$.  For an undirected graph $\cG$,  $\bM$ is symmetric and $\rkdual = r_k$. However, in general, this is not the case, but we assume that the influence is up to a certain degree mutual.
 In other words, we assume that if a node is very influential, it also is subject to the influence of many other nodes. We  make this precise in Section~\ref{sec:BARE}. 

As the performance measure, we compare any \emph{adaptive strategy} for this setting with the optimal oracle that knows $\bM$. The oracle strategy always chooses one of the most influential nodes, which are the nodes whose expected number of influences $r_k$ is maximal. We call one of these node $k^\star$, such that
\[k^\star = \argmax_k \EE{\sum_{t = 1}^n \left|S_{k,t}\right|} =  \argmax_k nr_k.\]
Let the reward of this node be 
\[r_\star = r_{k^\star}.\]
Then, its expected performance, if it consistently sampled $k^\star$ over $n$ rounds, is equal to
\[\EE{L^\star_n} = n r_\star.\]
The expected \emph{regret} of any adaptive strategy that is unaware of $\bM$, with respect to the  oracle strategy, is defined as the expected difference of the two, 
\[\EE{R_n} = \EE{L^\star_n} - \EE{L_n}.\]
Dually, we define $\rstardual$ as the average number of influences received by the most influenced node,
\[\rstardual = \max_{k}\rkdual.\]

\subsection{Baseline comparison: Observing only $\left|S_{k_t}\right|$, the number of influenced nodes}
\label{ss:first}
For a meaningful baseline comparison that shows the benefit of the graph structure, we first consider a \emph{restricted version} of the setting from Section~\ref{ss:prob}. The restriction is that the learner, at round $t$, does not observe the set of influenced nodes $S_{k_t,t}$, but only the number number of elements in $S_{k_t,t}$, denoted by $\left|S_{k_t,t}\right|$. In other words, once we select a node, we receive as a feedback only the number of influenced nodes, \emph{but not their identity}.
In this setting, we do not observe enough information about the graph structure to exploit it, since we do not observe the \textit{links} between the nodes. As a result, this setting can be mapped to a classic \emph{multi-arm bandit} setting without underlying graph structure, where the reward that the learner observes for node $k_t$ is equal to $\left|S_{k_t,t}\right|$.


If $n \geq d$, it is possible to directly apply classic multi-arm bandit reasoning. 
Since we never receive any information about the graph structure, we cannot exploit it and we can only consider the quantity $\left|S_{k_t,t}\right|$ as the standard bandit reward, which is a noisy version of $r_{k_t}$. Such  problem is a standard bandit problem with rewards  
$\left|S_{k_t,t}\right|$, that are integers between $0$ and $d$ and have a variance bounded by $r_{k_t}$. 

Directly building on upper and lower bounds arguments for the classic bandit strategies~\citep{lai1985asymptotically, audibert2009minimax}, we give the following result. This result's upper bound holds for a specific bandit algorithm that we call \GraphMOSS, a slight adaptation of the \MOSS algorithm by~\cite{audibert2009minimax} to our specific setting.

\begin{theorem}[proof in Appendix~\ref{proof:thls}]\label{thls}
In the graph bandit problem from Section~\ref{ss:first}, with the reward equal to the number of influenced nodes $\left|S_{k_t,t}\right|$ instead of $S_{k_t,t}$, the regret is bounded as follows.
\begin{itemize}[leftmargin=2.5em]
\item {\rm Lower bound.} If for some fixed $\varepsilon >0$, we have $\varepsilon d < r_\star < (1-\varepsilon)d$, then there exists a constant $\upsilon >0$ such that for $n$ large enough, depending on $\varepsilon$, we have that
\[\inf\sup\EE{R_n} \geq \upsilon \min\left( r_\star n, r_\star d+\sqrt{r_{\star}nd}\right),\]
where $\inf\sup$ means the best possible algorithm on the worst possible graph bandit problem.
\item {\rm Upper bound.}  There exists a constant $U>0$ such that the regret of Algorithm~\ref{alg:2} is bounded as
\[\EE {R_n} \leq U \min\left( r_\star n, r_\star d + \sqrt{r_{\star}nd}\right).\]
\end{itemize}
\end{theorem}

\begin{algorithm}[H]
\caption{\GraphMOSS}
 \label{alg:2}
 \begin{algorithmic}
 \STATE {\bfseries Input}
 \STATE \quad  $d$: the number of nodes
 \STATE \quad $n$: time horizon
  \STATE {\bfseries Initialization}
\STATE \quad Sample  each arm twice 
\STATE \quad  Update  $\widehat r_{k,2d}$,  $\widehat \sigma_{k,2d}$,
 and $T_{k,2d} \gets 2,$ for $\forall k\leq d$
\FOR{$t =2d+1, \dots, n$}
\STATE 
$C_{k,t} \gets 2 \widehat \sigma_{k,t} \sqrt{\frac{\max\left(\log(n/(dT_{k,t})),0\right)}{T_{k,t}}} $ 
\\ \hspace{5em} $+\ \frac{2\max(\log(n/(dT_{k,t})),0)}{T_{k,t}}$, for $\forall k\leq d$

\STATE   $k_t \gets \argmax_k \widehat r_{k,t} + C_{k,t}$
\STATE  Sample node $k_t$ and receive $|S_{k_t,t}|$
\STATE  Update  $\widehat r_{k,t+1}$,  $\widehat \sigma_{k,t+1}$, and  $T_{k,t+1}$, for $\forall k\leq d$
\ENDFOR
 \end{algorithmic}
\end{algorithm}

The lower bound holds also in the specific case where the graph $\cG$ is undirected (i.e.,\@ symmetric $\bM$), as is explained in the proof. This is an important remark  as the undirected graphs are a canonical and ``perfect" example of graphs where influencing and being influenced is correlated and where the dual influence is equal to the influence for each node.

\section{The  \BARE algorithm and results}
\label{sec:BARE}
In this section we treat the \emph{unrestricted} setting described in Section~\ref{ss:prob} where we \emph{get revealed the identity of the influenced nodes},  while the reward stays the same as in Section~\ref{ss:first}.  
First, note that the minimax-optimal rate in this setting is the same as in the restricted information case above. To see that, one can, for instance, consider a network composed of isolated nodes with only a very small clique of most influential nodes, connected only to each other.  Another example is a graph where the fact of being influential is uncorrelated with the fact of being influenced and where, for instance, the most influential node is not influenced by any node. For the same reasons as the ones described in Theorem~\ref{thls}, when $n \leq d$,  there is no adaptive strategy in a minimax sense, also in this unrestricted setting.

However, the cases where the identity of the influenced nodes does not help, are somewhat pathological. Intuitively, they correspond to cases where the graph structure is not very informative for finding the most influential node. This is the case when there are many isolated nodes, and also in the case where observing nodes that are very influenced does not provide information on these nodes' influence. In many typical and more interesting situations, this is not the case. First, in these problems, the nodes that have high influence are also very likely to be subject being influenced, for instance, many interesting networks are symmetric and then it is immediately the case. Second, in the realistic graphs, there is typically a small portion of the nodes that are noticeably more connected than the others~\citep{barabasi1999emergence}.

In order to rigorously define these non-degenerate cases, let us first define  function $D$ that controls the number of nodes with a given \emph{dual gap}, i.e.,~a given suboptimality with respect to the most influenced node 
\[ D(\Delta) \eqdef \left|\left\{i \leq d : \rstardual - \ridual \leq \Delta\right\}\right|. \]
The function $D(\Delta)$  is a non-decreasing quantity dual to the arm gaps. Note that $D(r) = d$ for any $r \geq \rstardual$ and that $D(0)$ is the number of most influenced nodes.
We now define the \emph{problem dependent} quantities that express the  difficulty of the problem and allow us to state our results.   
\begin{definition} 
 We define  the \textbf{detectable horizon} as the smallest integer $T_\star>0$ such that
\[T_\star\rstardual \geq \sqrt{D_\star n \rstardual},\]
when such $T_\star$ exists and  $T_\star = n$ otherwise. Here, $D_\star$ is  the \textbf{detectable dimension} defined as
\[D_\star \eqdef D(\Delta_\star),\]
where the \textbf{detectable gap} $\Delta_\star$ is defined as 
\[\Delta_\star \eqdef 16\sqrt{\frac{\rstardual d \log\left(nd\right)}{T_\star}}+ \frac{144d\log\left(nd\right)}{T_\star}\cdot\]
\end{definition}
\begin{remark} 
From the definitions above, the detectable dimension is the $D_\star$ that corresponds to the smallest integer $T_\star>1$ such that
\[T_\star\rstardual \geq \sqrt{D\left(16\sqrt{\frac{\rstardual d \log\left(nd\right)}{T_\star}}+ \frac{144d\log\left(nd\right)}{T_\star}\right) n \rstardual},\]
or $D_\star=d$ if such $T_\star$ does not exist. It is therefore a well defined quantity. Moreover, since $D$ is nondecreasing and $D(0)$ is the number of most influenced nodes, then $D_\star$ converges to the number of most influenced nodes as $n$ tends to infinity. 
\end{remark}

Finally let us write the influential-influenced gap as
$$\varepsilon_\star \eqdef r_\star - \max_{k\in \Ddualset} r_k,$$
where $\Ddualset \eqdef \{ i : \ridual  = \max_k \rkdual\}$. 
The quantity $\varepsilon_\star$ quantifies the gap between the most influential node overall vs.\@ the most influential node in the set of most influenced nodes.
\begin{remark} 
The quantity $\varepsilon_\star$ is small when one of the most influenced nodes is also very influential.
It is exactly zero when one of the most influential nodes happens to also be one of the most influenced nodes. For instance, the case   $\varepsilon_\star = 0$  appears in undirected social network models with mutual influence. 
\end{remark} 

The graph structure is helpful when the $D$ function decreases quickly with $n$. 
To give an intuition about how is $D$ linked to the graph topology, consider  a star-shaped graph which is  the most helpful and can have $D_\star = 1$ even for a small $n$. On the other hand, a bad case is a graph with many small cliques. The worst case is where all nodes are disconnected except two, where $D_\star$ will be of order $d$ even for a large $n$. 

The detectable dimension $D_\star$ is a problem dependent quantity  that represents the \emph{complexity of the problem} instead of $d$.  In real networks, $D_\star$ is typically smaller than the number of nodes $d$ and we give several examples of the empirical value of $D_\star$ in Section~\ref{sec:exp} and Appendix~\ref{app:graphs}.
 As our analysis will show, $D_\star$ represents the number of nodes that we can \emph{efficiently extract} from~$d$ nodes in less than $n$ rounds of the time budget. Our \emph{bandit revelator} algorithm, \BARE (Algorithm~\ref{alg:1}), starts by the \emph{global-exploration} phase and extracts a subset of cardinality less than or equal to a constant multiple of $D_\star$, that contains a very influential node, that is at most $\varepsilon_\star$ away from the most influential node. \BARE does this extraction  \textit{without scanning all the $d$ nodes}, which could be impossible anyway, since we do not restrict to $d \leq n$.
In the subsequent \emph{bandit} phase, \BARE proceeds with scanning this smaller set of selected nodes to find the most influential one.

\begin{algorithm}[t]
\caption{\BARE: \underline{Ba}ndit \underline{re}velator}
 \label{alg:1}
 \begin{algorithmic}
 \STATE {\bfseries Input}
 \STATE \quad  $d$: the number of nodes
 \STATE  \quad $n$: time horizon
  \STATE {\bfseries Initialization}
 \STATE \quad $T_{k,t} \gets 0,$ for $\forall k\leq d$
\STATE \quad 
$ \widehat \rktdual \gets 0$, for $\forall k\leq d$
\STATE \quad $t \gets 1$,
$\widehat T_{\star} \gets 0$,
$\widehat D_{\star,t} \gets d$,
$\widehat \sigma_{\star,1} \gets d$
\STATE {\bfseries Global exploration phase}
\WHILE{$t \left(\widehat \sigma_{\star,t} - 4\sqrt{d\log(dn)/t}\right) \leq \sqrt{\widehat D_{\star,t} n }$}
\STATE Influence a node at random (choose $k_t$ uniformly at random) and get $S_{k_t,t}$ from this node
\STATE  $\widehat \rkttdual \gets \frac{t}{t+1}\widehat \rktdual + \frac{d}{t+1} S_{k_t,t}(k)$
\STATE  $\widehat \sigma_{\star,t+1} \gets \max_{k'}\sqrt{\widehat \rkprimettdual + 8d\log(nd)/(t+1) } $
\STATE $w_{\star,t+1} \gets 8\widehat \sigma_{\star,t+1}\sqrt{\frac{d \log\left(nd\right)}{t+1}} + \frac{40d\log\left(nd\right)}{t+1}$
\STATE  $\widehat D_{\star,t+1} \gets \left|\left\{k : \max_{k'} \widehat \rkprimettdual - \widehat \rkttdual \leq  
w_{\star,t+1} \right\}\right|$
\STATE  $t \gets t+1$
\ENDWHILE
\STATE  $\widehat T_\star \gets  t$.
\STATE {\bfseries Bandit phase}
\STATE Run minimax-optimal bandit algorithm  on the $\widehat D_{\star,\widehat T_\star}$ chosen nodes
(e.g., Algorithm~\ref{alg:2})
 \end{algorithmic}
\end{algorithm}

We now state our main theoretical result that proves a bound on the regret of \BARE. 

\begin{theorem}[proof in Section~\ref{p:BARE}] \label{BARE}
In the unrestricted local influence setting with information about the neighbors, \BARE satisfies, for a constant $C>0$,
\[\EE{R_n} \leq C\min\left(r_\star n, D_\star r_{\star} + \sqrt{r_{\star}nD_\star} + n \varepsilon_\star\right).\]

\end{theorem}

\begin{remark}
Note that \BARE does not need preliminary information about $\cG$, as a classic multi-arm bandit strategy described in Section~\ref{ss:first} would require in order to attain this rate. 
\end{remark}

\begin{corollary}

For an undirected social network model the  expected regret of \BARE is
\[\EE{R_n} \leq C\min\left(r_\star n, D_\star r_{\star} + \sqrt{r_{\star}nD_\star}\right),\]
which is the minimax-optimal regret in the case where there are $D_\star$ instead of $d$ nodes. 
This highlights the dimensionality reduction potential of our method.
\end{corollary}

Finally,  we state a lower bound for our setting. Notice that the influential-influence gap also appears here.

\begin{theorem}[proof in Appendix~\ref{p:BARE2}] \label{BARE2}
Let $d \geq Cn>0$, where $C>0$ is a universal constant. Consider the set of unrestricted settings and the set of all problems that have maximal influence bounded by $r$ (where for some fixed $u >0$, we have $u \overline D < r < (1-u)\overline D$), $D_\star \leq \overline D$ and influential-influence gap smaller than $\varepsilon$ (with $\varepsilon\leq u\sqrt{dr/n}$ for some small $u>0$ if $d\leq n$). Then the expected regret of the best possible algorithm in the worst case of these problems is lower bounded as
\[C''\min\left(rn,  \overline Dr + \sqrt{rn\overline D}  + n \varepsilon\right),\]
where $C''$ is a universal constant.
\end{theorem}

\section{Proof of Theorem~\ref{BARE}}\label{p:BARE}

For any node $k \leq d$ and any round $t$ that is during the \emph{global exploration phase}, 
let us define the following estimator of reward $\rktdual$,
\[\widehat \rktdual = \frac{1}{t}\sum_{t'=1}^t dS_{k_t,t'}(k) .\]
Notice that during the global exploration phase, the nodes are chosen uniformly at random among all the nodes.
This means that for any $k$, the $(S_{k_t,t'}(k))_{t'}$ are i.i.d.\@ Bernoulli random variables with parameter $\rkdual/d$. By Bernstein inequality, this implies that with probability larger than $1-1/n^2$, for any node $k \leq d$ and for any round $t$ within the global exploration phase, 
\begin{equation}\label{eq:xi}
\left|\widehat \rktdual - \rkdual\right| \leq 4\sqrt{\frac{d \rkdual\log\left(nd\right)}{t} } + \frac{4d \log\left(nd\right)}{t}\cdot
\end{equation}
Let $\xi$ be the event such that Equation~\ref{eq:xi} holds. Note that on $\xi$, we have that for any $t$ of the global exploration phase and for any $k \leq d$,
\begin{align*}
\rkdual - 4\sqrt{\frac{d \rkdual\log\left(nd\right)}{t}} +\frac{4d\log\left(nd\right)}{t} &\leq \widehat \rktdual + \frac{8d\log\left(nd\right)}{t} \\
&\hspace{-8em}\leq \rkdual + 4\sqrt{\frac{d \rkdual\log\left(nd\right)}{t}} + \frac{12d\log\left(nd\right)}{t},
\end{align*}
which implies that on $\xi$, by factorizing the left hand side and right hand side, for any $k \leq d$,
\begin{align*}
\left(\sqrt{\rkdual} - 2\sqrt{\frac{d\log\left(nd\right)}{t}}\right)^2 &\leq \widehat \rktdual + \frac{8d\log\left(nd\right)}{t} \\ &\leq \left(\sqrt{\rkdual} + 4 \sqrt{\frac{d\log\left(nd\right)}{t}}\right)^2\!\!\!,
\end{align*}
which implies
\begin{align*}
\sqrt{\rkdual} - 2\sqrt{\frac{d\log\left(nd\right)}{t}} &\leq \sqrt{\widehat \rktdual + \frac{8d\log\left(nd\right)}{t}} \\ &\leq \sqrt{\rkdual} + 4 \sqrt{\frac{d\log\left(nd\right)}{t}}\cdot
\end{align*}
We deduce from this that
\[\left|\sqrt{\rkdual} - \sqrt{\widehat \rktdual + \frac{8d\log\left(nd\right)}{t}}\right| \leq 4 \sqrt{\frac{d\log\left(nd\right)}{t}}\cdot\]
In particular, this implies that on $\xi$, 
\begin{equation}\label{eq:sig}
\left|\widehat \sigma_{\star,t} - \sqrt{\rstardual}\right| \leq 4\sqrt{\frac{d\log\left(nd\right)}{t}}\cdot
\end{equation}
On $\xi$, we also have by Equation~\ref{eq:xi},
\begin{align*}
&\left|\left(\max_{k'}\widehat \rkprimetdual -  \widehat \rktdual\right) - \left(\rstardual - \rkdual\right)\right| \\
&\hskip 2cm \leq 8\sqrt{\frac{d \rstardual\log\left(nd\right)}{t}} + \frac{8d\log\left(nd\right)}{t},
\end{align*}
which implies that on $\xi$, by Equation~\ref{eq:sig},
\begin{align*}
&\left|\left(\max_{k'}\widehat \rkprimetdual -  \widehat \rktdual\right) - \left(\rstardual - \rkdual\right)\right| \\
&\hskip 2cm \leq 8\widehat \sigma_{\star,t} \sqrt{\frac{d \log\left(nd\right)}{t}} + \frac{40d\log\left(nd\right)}{t}\cdot
\end{align*}

Note that by the definition of the global exploration phase, we know that for any round $ t \leq \widehat T_\star$, the set of most influenced nodes $\Ddualset$ will be 
on $\xi$
 in the set of the $\widehat D_{\star, t}$ kept nodes. Note that by Equation~\ref{eq:sig}, this also implies that on $\xi$,
\begin{align}\label{eq:D}
\widehat D_{\star,t} &\leq D\left(16 \widehat \sigma_{\star,t} \sqrt{\frac{d  \log\left(nd\right)}{t}} + \frac{80d\log\left(nd\right)}{t}\right) \nonumber \\
&\leq D\left(16\sqrt{\frac{d \rstardual \log\left(nd\right)}{t}} + \frac{144d\log\left(nd\right)}{t}\right)\cdot
\end{align}



\paragraph{First case: the global exploration phase finishes before $3T_\star$}
We consider the case $\widehat T_\star \leq 3T_\star$. If the exploration finishes at $\widehat T_\star$, then on $\xi$, by Equation~\ref{eq:sig}, and by the definition of \BARE, 
\[3T_\star \sqrt{\rstardual} \geq \widehat T_\star \sqrt{\rstardual} \geq \sqrt{\widehat D_{\star,\widehat T_\star} n }.\]
By the definition of $D_\star$ we also have that
\[ \sqrt{D_\star n \rstardual} \geq (T_\star-1)\rstardual \geq T_\star\rstardual/2,\]
which together implies
\[\widehat D_{\star,\widehat T_\star} \leq 36D_\star.\]
Also, on $\xi$, the optimal arm is among the $\widehat D_{\star,\widehat T_\star}$  arms.

%
\paragraph{Second case: the global exploration phase finishes after $3T_\star$}
The detectable gap $\Delta_\star$ is equal to
\[\Delta_\star =  16\sqrt{\frac{\rstardual d \log\left(nd\right)}{T_\star}}+ \frac{144d\log\left(nd\right)}{T_\star}\cdot\]
Since the detectable dimension $D_\star$ is smaller or equal to $d$, then $\Delta_\star \leq \rstardual$. This implies that $T_\star$ must satisfy
\[\rstardual \geq  16\sqrt{\frac{\rstardual d \log\left(nd\right)}{T_\star}}+ \frac{144d\log\left(nd\right)}{T_\star},\]
which implies that
\begin{equation}\label{eq:lbT}
T_\star \geq  \frac{144d\log\left(nd\right)}{\rstardual}\cdot
\end{equation}
In the case we consider $\left(\widehat T_\star \geq 3T_\star\right)$, the exploration phase does not stop at $3T^\star$ and we have that on~$\xi$,
\[3T_\star \left(\widehat \sigma_{\star,T^\star} - 4 \sqrt{\frac{d\log\left(nd\right)}{T_\star}}\right) \leq \sqrt{\widehat D_{\star,  T_\star} n },\]
and thus by Equation~\ref{eq:sig}, we have that
\[3T_\star \left(\sqrt{\rstardual} - 8 \sqrt{\frac{d\log\left(nd\right)}{T_\star}}\right) \leq \sqrt{\widehat D_{\star,  T_\star} n },\]
which implies in turn by Equation~\ref{eq:lbT} that
\[\frac{3T_\star \sqrt{\rstardual}}{3} \leq \sqrt{\widehat D_{\star, T_\star} n }.\]
Combining Equation~\ref{eq:D}, with the fact that $D$ is a nondecreasing function, we get that on~$\xi$,
\begin{align*}
T_\star \sqrt{\rstardual} &\leq \sqrt{D\left(16\sqrt{\frac{d \rstardual\log\left(nd\right)}{T_\star}} + \frac{144d\log\left(nd\right)}{T_\star}\right) n } \\
&\leq \sqrt{D_\star n },
\end{align*}
which is false by definition of $D_\star$ and $T_\star$. Therefore, we know that on $\xi$, $3T_\star \geq \widehat T_\star$ .


\paragraph{Conclusion}
To sum up, we know that on $\xi$,
\[\widehat T_\star \leq 3T_\star \text{\quad and \quad} 
\widehat D_{\star, \widehat T_\star} \leq 36 D_\star,\]
and that the set of most influenced nodes $\Ddualset$ is among the nodes that are kept at the end of the global exploration phase. In particular, this implies that the gap with respect to the most influential node on this set is at most $\varepsilon_\star$.


Taking the $\widehat D_{\star, \widehat T_\star} \leq 36 D_\star$ kept arms and running a minimax bandit algorithm, such as \GraphMOSS, we can  upper bound the regret incurred in the remaining rounds using Theorem~\ref{thls}. Since there are $n- T_\star$ remaining rounds,  this  implies that the expected regret on these last rounds, on $\xi$, for a given constant $C'>0$, bounded by
\[C'D_\star r_\star + C'\!\sqrt{r_\star \widehat D_{\star, \widehat T_\star}\left(n\!-\!\widehat T_\star\right)} \leq C'D_\star r_\star+ C'\sqrt{r_\star D_{\star}n},\]
with respect to the optimal nodes in the set of kept nodes. 
Now, since $\Ddualset$ is in the set of kept nodes, and since the maximal gap of most influential nodes with respect to this set is at most $\varepsilon_\star$,  the regret with respect to the most influential node $r_\star$ is
\begin{align*}
C'D_\star r_\star +  &C'\sqrt{r_\star \widehat D_{\star, \widehat T_\star}\left(n - \widehat T_\star\right)} + n \varepsilon_\star \\ &\leq C'D_\star r_\star+ C'\sqrt{r_\star D_{\star}n} + n\varepsilon_\star.
\end{align*}
We can now conclude the proof by bounding the expected regret as
\begin{align*}
\EE{R_n} &\leq T_\star r_\star + C'D_\star r_\star+ C'\sqrt{r_\star D_{\star}n}+n\varepsilon_\star + \frac{r_\star}{n} \\
&\leq \left(C' +2\right)\sqrt{r_\star D_{\star}n} + 2C' r_\star D_\star + n\varepsilon_\star \\
&\leq C\left(\sqrt{r_\star D_{\star}n} + r_\star D_\star\right) + n\varepsilon_\star.
\end{align*}
%
%

\begin{figure*}[t]
\begin{center}
\vspace{-3.5em}
\begin{picture}(115,115)
\put(0,0){\includegraphics[width=0.495\columnwidth]{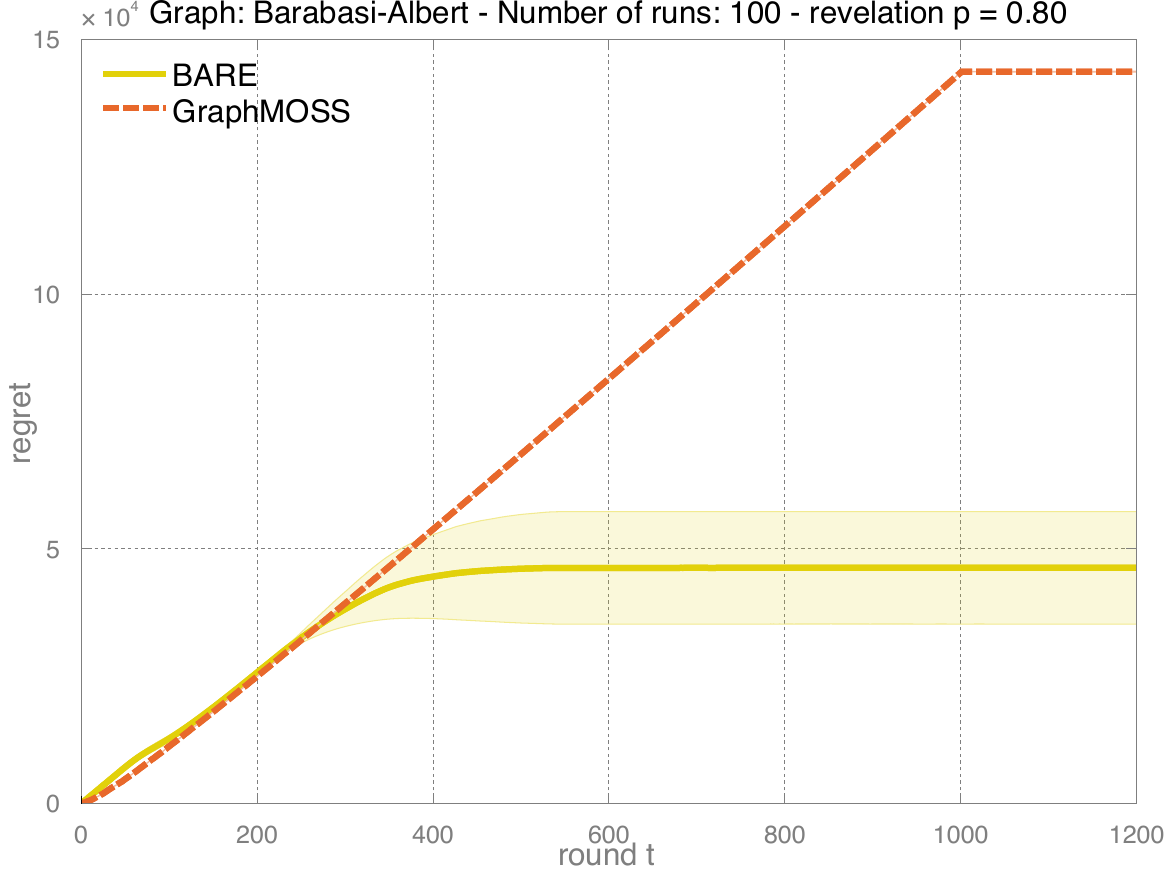}}
\put(10,60){\tiny $\widehat D_{\star} = 134$, $\widehat T_{\star} = 36$}
\end{picture}
\begin{picture}(115,115)
\put(0,0){\includegraphics[width=0.495\columnwidth]{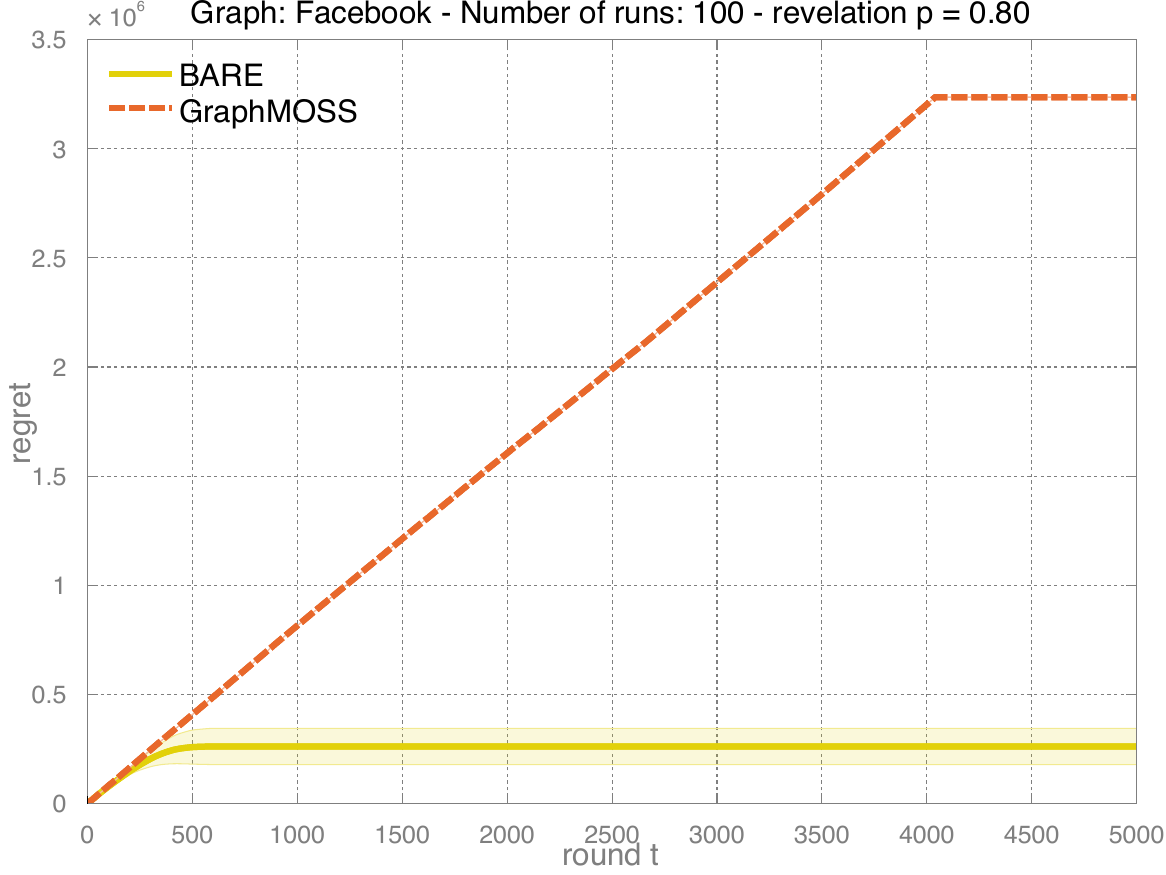}}
\put(10,60){\tiny $\widehat D_{\star} = 125$, $\widehat T_{\star} = 28$}
\end{picture}
\begin{picture}(115,115)
\put(0,0){\includegraphics[width=0.495\columnwidth]{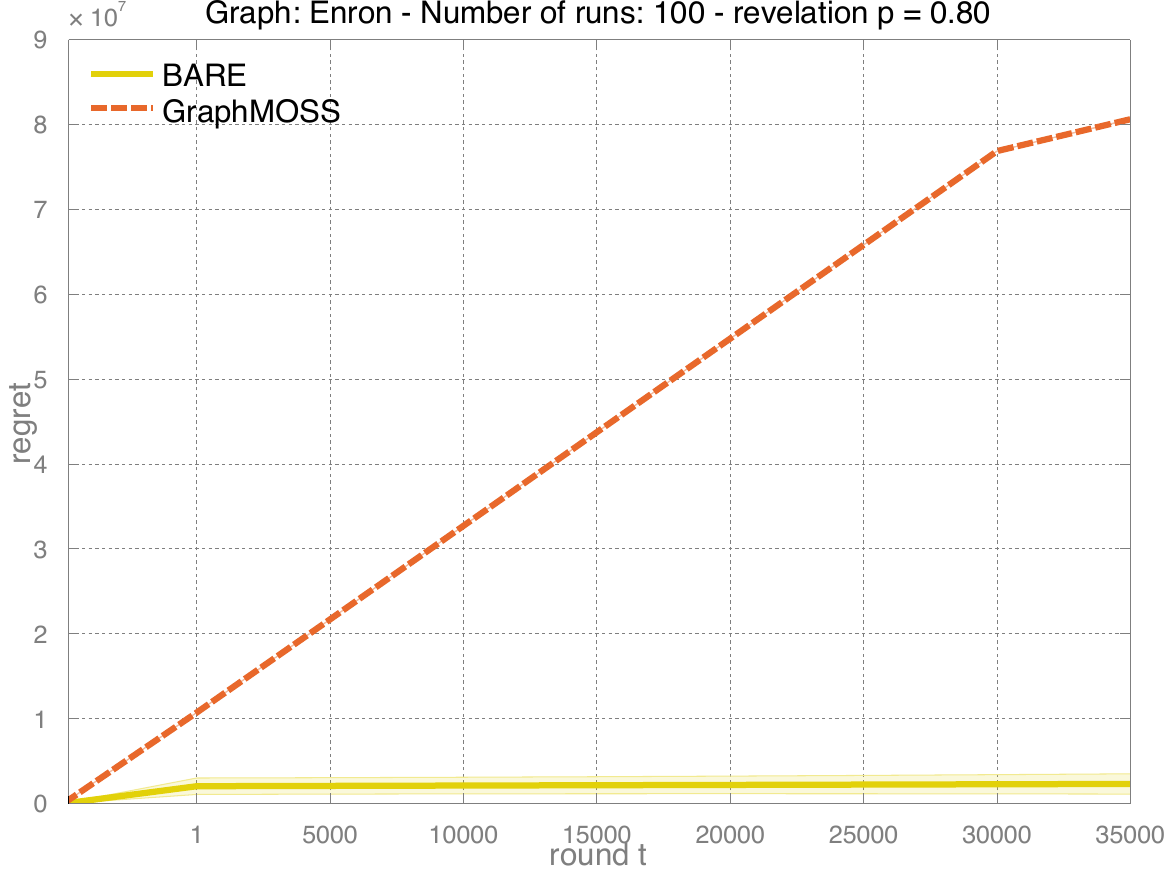}}
\put(10,60){\tiny $\widehat D_{\star} = 564$, $\widehat T_{\star} = 107$}
\end{picture}
\begin{picture}(115,115)
\put(0,0){\includegraphics[width=0.495\columnwidth]{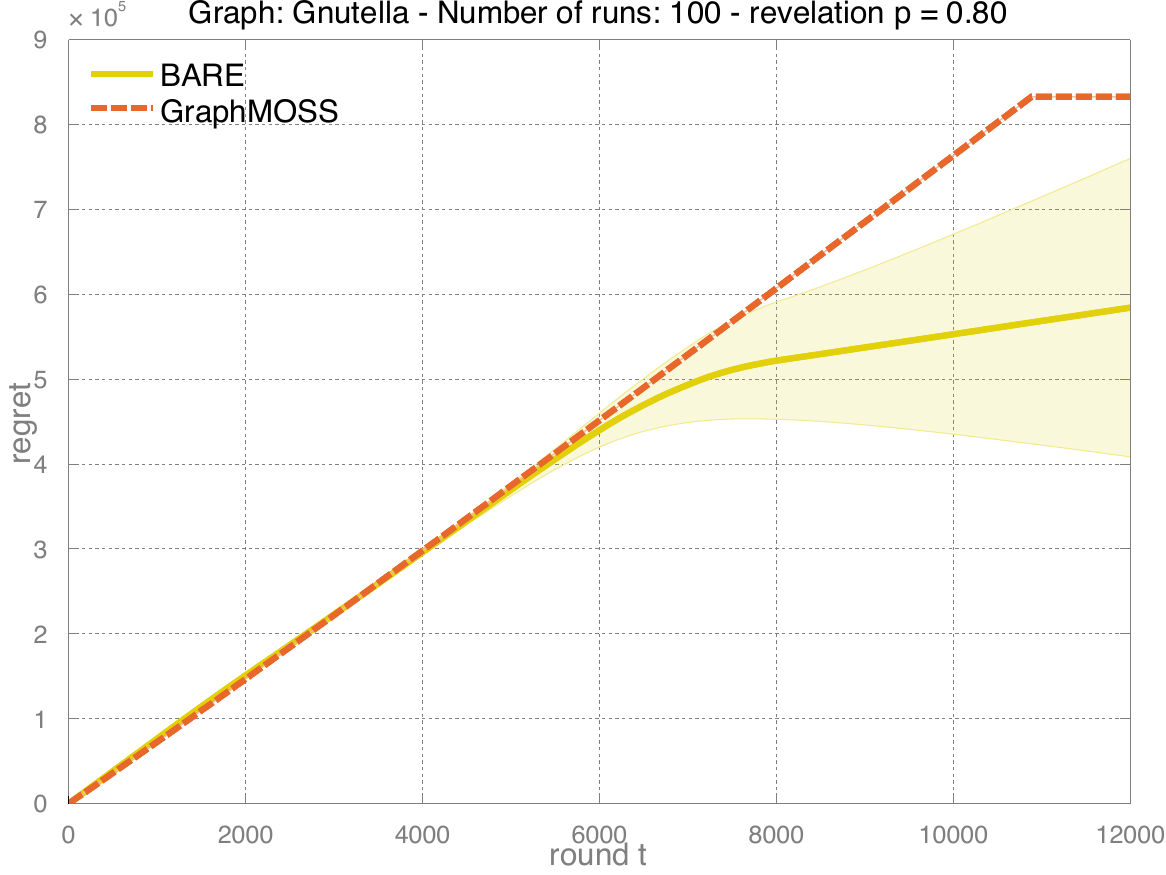}}
\put(10,60){\tiny $\widehat D_{\star} = 3916$, $\widehat T_{\star} = 779$}
\end{picture}
 \vspace{-1.0em}
\caption{%
\textit{Left}: Barab\'asi-Albert. 
\textit{Middle left}: Facebook. 
\textit{Middle right}: Enron. 
\textit{Right}: Gnutella.}
\label{fig:ba}
\end{center}
\end{figure*}

\begin{figure*}[t]
\begin{center}
\vspace{-2.0em}
\begin{picture}(90,90)
\put(0,0){\includegraphics[width=0.40\columnwidth]{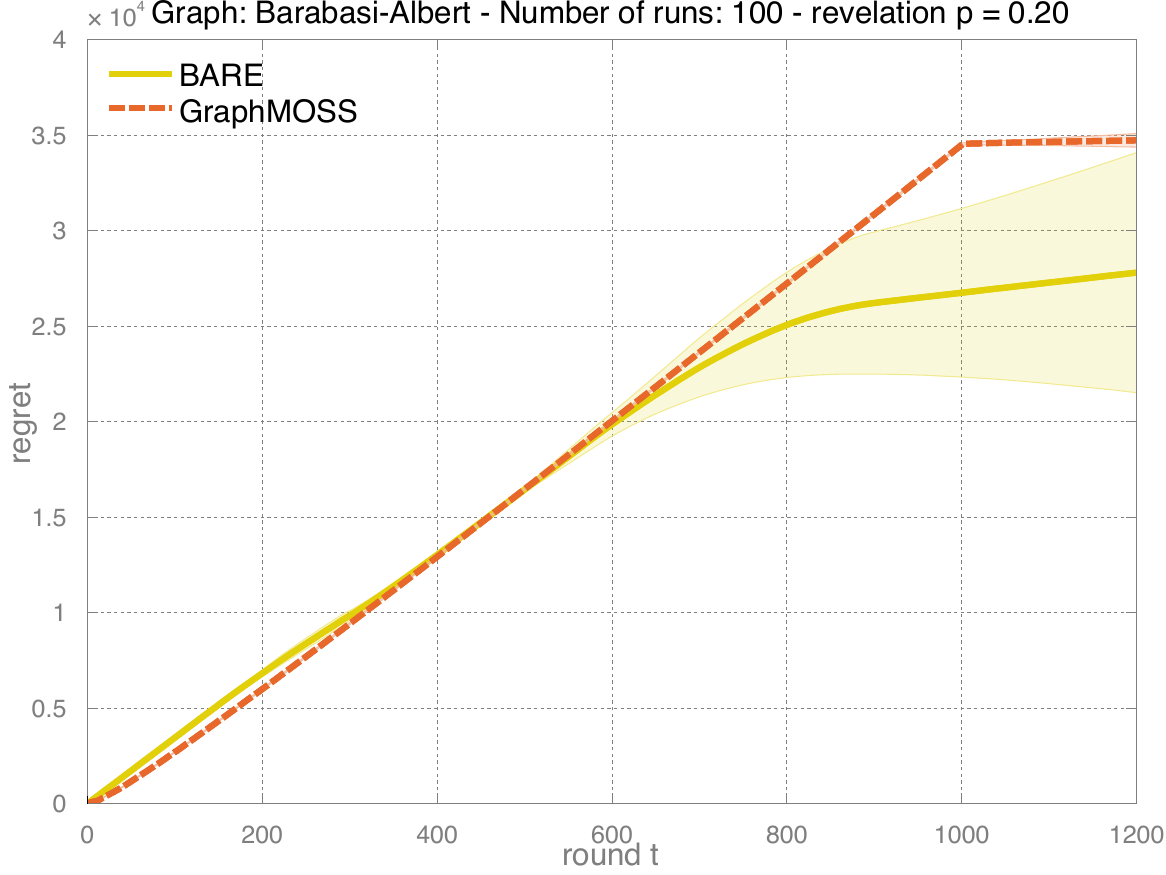}}
\put(10,50){\tiny $\widehat D_{\star} = 529$, $\widehat T_{\star} = 147$}
\end{picture}
\begin{picture}(90,90)
\put(0,0){\includegraphics[width=0.40\columnwidth]{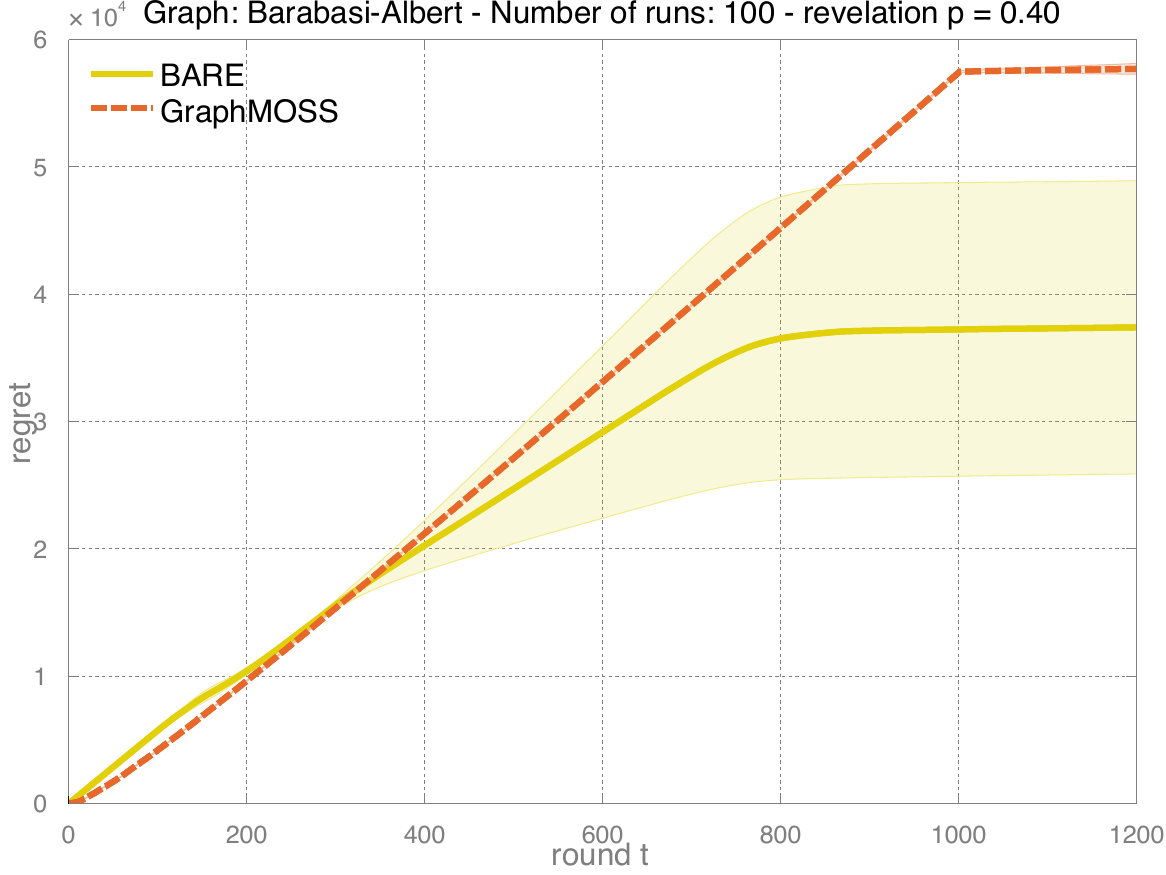}}
\put(10,50){\tiny $\widehat D_{\star} = 230$, $\widehat T_{\star} = 64$}
\end{picture}
\begin{picture}(90,90)
\put(0,0){\includegraphics[width=0.40\columnwidth]{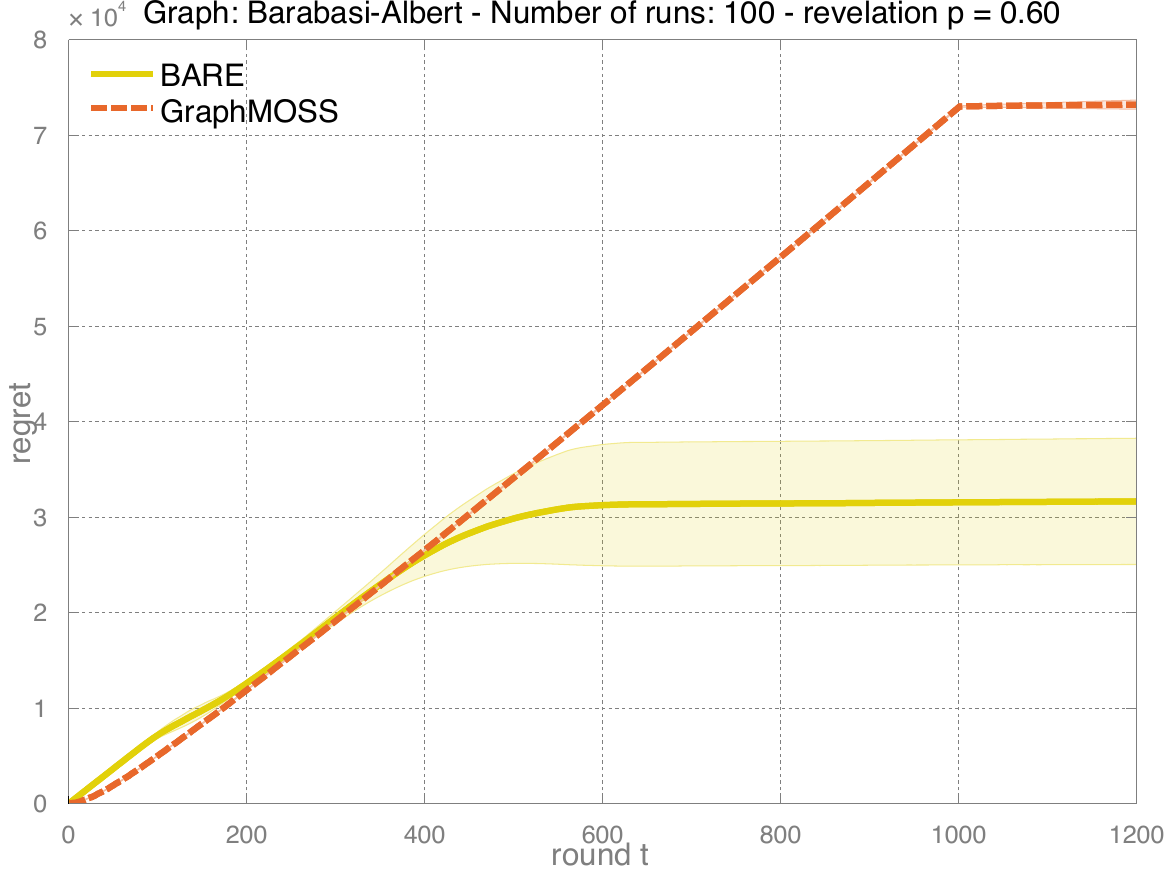}}
\put(10,50){\tiny $\widehat D_{\star} = 161$, $\widehat T_{\star} = 50$}
\end{picture}
\begin{picture}(90,90)
\put(0,0){\includegraphics[width=0.40\columnwidth]{fig/v2ba-1000-0_80}}
\put(10,50){\tiny $\widehat D_{\star} = 134$, $\widehat T_{\star} = 36 $}
\end{picture}
\begin{picture}(90,90)
\put(0,0){\includegraphics[width=0.40\columnwidth]{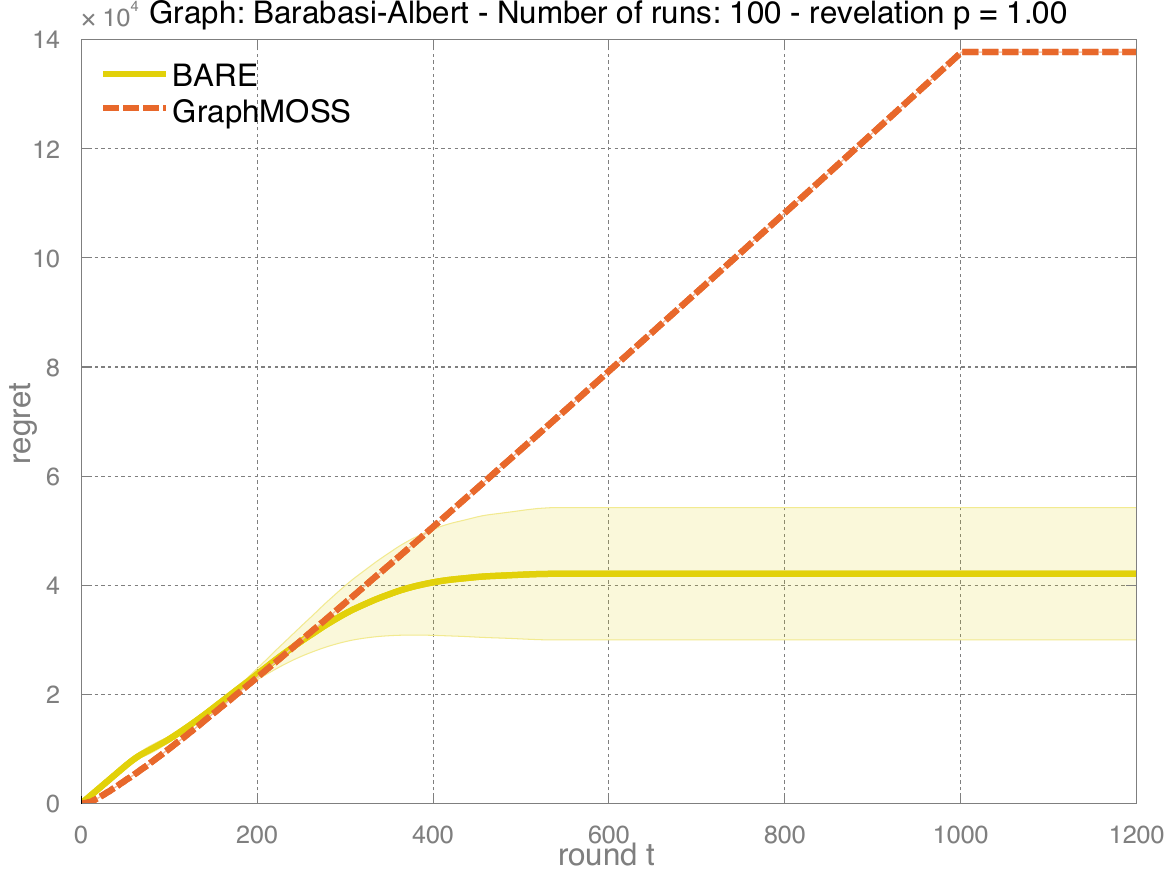}}
\put(10,50){\tiny $\widehat D_{\star} = 133$, $\widehat T_{\star} = 34 $}
\end{picture}
 \vspace{-1.0em}
\caption{Barab\'asi-Albert model with varying $p$ between 0.2 and 1}
\label{fig:p}
\end{center}
\end{figure*}
 
\subsection{Discussion}
 
\paragraph{Lower bound} Theorem~\ref{BARE2} holds in the case $Cn \leq d$ and makes the quantity $\varepsilon_\star$ appear. But we emphasize that in the case $n \geq d$,  even if the oracle provides the learner with a set of nodes such that the optimal node belongs to this set, the minimax-optimal rate for the bandit problem becomes  
\[C\min\left(r_\star n, D_\star r_\star + \sqrt{r_{\star}nD_\star}\right),\] for a constant  $C$.
This can be seen from an argument similar to Theorem~\ref{thls}, together with the example with isolated nodes, given above. 
This argument holds even for undirected graphs with $\varepsilon_\star = 0$. In this sense, \BARE is minimax-optimal over the set of problems with detectable dimension $D_\star$.

\paragraph{Large scale setting} The quantity $D_\star$ and \BARE become particularly appealing when we consider an interesting practical situation with a large number of graph nodes. For instance, even in a medium-sized social network, the advertiser 
would not have enough budget to target all the users and discover the most influential one, i.e.,\@ $n \leq d$. 
Notice again, that in the restricted setting of Section~\ref{ss:first}, the regret of bandit strategies in this problem for $n \ll d$ is of order
$n r_\star,$
which is larger than the regret of \BARE.

However, in the unrestricted setting, the situation is different when $D_\star \leq n$. This is the case  where a small number of nodes is noticeably more influential than the others and the regret of \BARE is of order
\[D_\star r_{\star} + \sqrt{r_{\star}nD_\star} + n\varepsilon_\star,\]
which is smaller than $n r_\star$, and the problem becomes \emph{learnable}.

\section{Experiments}
\label{sec:exp}
The purpose of our experiments is to show that \BARE can do better in the 
regime $n \leq d$, compared to the algorithms ignoring the graph structure. 
For the minimax optimal algorithm during the bandit phase of \BARE, we used \GraphMOSS, defined  
in Section~\ref{ss:first} and analyzed in Appendix~\ref{proof:thls}, which is a close  variation of the \MOSS algorithm~\citep{audibert2009minimax}. 
We also used \GraphMOSS as the baseline algorithm that does not use the graph structure.

The confidence parameter $\delta$ was set to 0.01 and  $p_{i,j}$ to 0.8 for all $i$ and $j$. 
This means that whenever a node is chosen, each of its neighbors is influenced and revealed with probability 0.8.
Since the confidence terms of \BARE are conservative, in the experiments we multiplied them by 0.01.
All figures show the results averaged over 100 trials.

We first performed an experiment on a graph generated by 10-out-degree Barab\'asi-Albert model with $d = 1000$ nodes. Figure~\ref{fig:ba} (left) compares \BARE with \GraphMOSS.
 As expected, \GraphMOSS suffers linear regret up to time $t = d$, since there is no sharing of information and for $t\leq d$, \GraphMOSS pulls each arm once. 
 While the regret of \GraphMOSS is no longer linear for $t > d$ and eventually detects the best node, \BARE is able to detect promising nodes much sooner during its global exploration phase and we can see the benefit of revealed information already around $t = 300$. 
 
In Figure~\ref{fig:p}, we varied the probability of revelation~$p$ for a Barab\'asi-Albert graph. When $p$ close is to one, the more of the graph structure
is revealed and the problem becomes easier. On the other hand, with~$p$ close to zero we do not get as much information about the structure and the performance 
of \BARE and \GraphMOSS are similar. 
 
%

We also performed the experiments on Enron mail graph~\citep{klimt2004introducing} with $d=36692$ and the snapshot of 
symmetrized version of Gnutella network from  August 4th, 2002 ~\citep{ripeanu2002mapping} with $d=10879$, obtained from Stanford Large Network Dataset Collection \citep{leskovec2014snapnets}. Furthermore, we evaluated \BARE on a subset of Facebook network with $d=4039$~\citep{viswanath2009evolution}.  We used the same parameters as for the 
Barab\'asi-Albert case. 

As expected, Figure~\ref{fig:ba} (middle left, middle right, right) shows that the performance gains of \BARE over \GraphMOSS depend heavily on the structure. 
In Enron and Facebook, the gain of \BARE is significant which suggests that the graphs from these networks feature a relatively small number of influential nodes. On the other hand, 
the gain of \BARE on Gnutella was much smaller which again suggests that this network is more decentralized.

In all the plots we include also the empirical estimate of the detectable dimension~$\widehat  D_{\star}$ and the detectable horizon~$\widehat T_{\star}$. Notice that the smaller $\widehat  D_{\star}$,  as compared to~$d$, and the smaller  $\widehat T_{\star}$ is as compared to~$n$, the sooner is \BARE able to learn the most influential node as compared to \GraphMOSS.


%
%
%

\section{Conclusion}

We hope that out work on local revelation incites the extensions on more elaborate propagation models on graphs~\citep{kempe2015maximizing}. One way to directly extend to more general propagation models is to consider that a more distant neighbor is a direct neighbor with contamination probability being the sum of the path products. Moreover, if we allow for more feedback, e.g., the identity of the influencing paths, our results could extend more efficiently.
Note that in our setting, we were completely agnostic to the graph structure. Realistic networks often exhibit some additional structural properties that are captured by graph generator models, such as various \emph{stochastic block models}~\citep{girvan2002community}. 
In future, we would like to extend our approach to cases where we can take advantage of the assumptions  stemming from these models and consider the subclasses of graph structures where we can further improve the learning rates.

\paragraph{Acknowledgements}
\label{sec:Acknowledgements} We thank Alan Mislove for the Facebook dataset.
The research presented in this paper was supported by French Ministry of
Higher Education and Research, Nord-Pas-de-Calais Regional Council,  French National Research Agency project ExTra-Learn (n.ANR-14-CE24-0010-01), 
and by German Research Foundation's Emmy Noether grant MuSyAD (CA 1488/1-1).

\vfil


{\fontsize{9.5}{11.5}\selectfont
\bibliography{library}
\bibliographystyle{icml2015}
}


\clearpage
\appendix


\onecolumn

\section{Proof of Theorem~\ref{thls}}\label{proof:thls}
\setcounter{theorem}{1-1}
\begin{theorem}
In the graph bandit problem from Section~\ref{ss:first}, with the reward equal to the number of influenced nodes $\left|S_{k_t,t}\right|$ instead of $S_{k_t,t}$, the regret is bounded as follows.
\begin{itemize}[leftmargin=2.5em]
\item {\rm Lower bound.} If for some fixed $\varepsilon >0$, we have $\varepsilon d < r_\star < (1-\varepsilon)d$, then there exists a constant $\upsilon >0$ such that for $n$ large enough, depending on $\varepsilon$, we have that
\[\inf\sup\EE{R_n} \geq \upsilon \min\left( r_\star n, r_\star d+\sqrt{r_{\star}nd}\right),\]
where $\inf\sup$ means the best possible algorithm on the worst possible graph bandit problem.
\item {\rm Upper bound.}  There exists a constant $U>0$ such that the regret of Algorithm~\ref{alg:2} is bounded as
\[\EE {R_n} \leq U \min\left( r_\star n, r_\star d + \sqrt{r_{\star}nd}\right).\]
\end{itemize}
\end{theorem}

The upper bound follows immediately from the results of~\cite{audibert2009minimax} using the Bernstein bound in the  confidence term in the \MOSS algorithm~\citep{audibert2009minimax}. The confidence term in the resulting 
Algorithm~\ref{alg:2}
 becomes
\begin{align*}
C_{k,t} =
2&\widehat \sigma_{k,t} \sqrt{\frac{\max\left(\log(n/(dT_{k,t})),0\right)}{T_{k,t}}}  + \frac{2\max(\log(n/(dT_{k,t})),0)}{T_{k,t}} \nonumber,
\end{align*}
where $T_{k,t}$ is the number of pulls of node $k$ at round $t$, and $\widehat \sigma_{k,t}^2$ is the empirical variance of node $k$ at round~$t$. This resulting scaling is then bounded by $r_\star$, since $\left|S_{k_t,t}\right|$ is a sum of Bernoulli random variables such that the sum of their variances is $\sigma_{k_t}^2 \leq r_{k_t} \leq r_\star$, which implies that the variance of $\left|S_{k_t,t}\right|$ is bounded by $r_\star$. The result is obtained following the scheme of~\cite{audibert2009minimax}.
 
The lower bound can be deduced from the one of~\cite{lai1985asymptotically}, replicating their $d$-armed problems with gap $\sqrt{d/n}$ by an equivalent problem with the graph structure where we define the following setup:
\begin{itemize} 
\item One arm influences any arm with probability $ r_\star/d$ with $1 - \varepsilon > r_\star/d >\varepsilon$.
\item The other arms influence each other with probability $r_\star/d - \sqrt{r_\star/(dn)}$ that is between $1 - \varepsilon$ and $\varepsilon/2$ for $n$ large enough, i.e., larger than $d$ times a universal constant.
\end{itemize}
The gap with respect to any suboptimal arm is~$\sqrt{dr_\star/n}$ and the variance for each arm is of order~$r_\star$ for $n$ large enough (larger than $d$ times a universal constant). 
The statement of the lower bound by the adaptation of the proof of~\cite{lai1985asymptotically} to this specific sub-Gaussian reward, i.e., sum of Bernoulli random variables with a parameter between $1 - \varepsilon$ and $\varepsilon/2$.


For both the upper and the lower bound, the quantity $r_\star d$ corresponds to the fact that each arm must be sampled at least once in the case of $n \geq d$ and the quantity $r_\star n$ corresponds to the \emph{unlearnable} case where $n \leq d$. Otherwise, one can always consider the worst case permutation of the arms.

\section{Proof of Theorem~\ref{BARE2}}\label{p:BARE2}

\setcounter{theorem}{3-1}
\begin{theorem}
Let $d \geq Cn>0$, where $C>0$ is a universal constant. Consider the set of unrestricted settings and the set of all problems that have maximal influence bounded by $r$ (where for some fixed $u >0$, we have $u \overline D < r < (1-u)\overline D$), $D_\star \leq \overline D$ and influential-influence gap smaller than $\varepsilon$ (with $\varepsilon\leq u\sqrt{dr/n}$ for some small $u>0$ if $d\leq n$). Then the expected regret of the best possible algorithm in the worst case of these problems is lower bounded as
\[C''\min\left(rn,  \overline Dr + \sqrt{rn\overline D}  + n \varepsilon\right),\]
where $C''$ is a universal constant.
\end{theorem}

If $\sqrt{nr\overline D} \geq a n\varepsilon$ for a small $a>0$, we consider the following construction. We consider the construction of the lower bound in Theorem~\ref{thls}, where the player also receives the position of the $\overline D$ best nodes as an additional information. This situation is therefore easier than the full problem. Let $\mathcal S$ be the set of $\overline D$ best nodes. For any $k, l \in \mathcal S^2$, if neither $k$ or $l$ is the optimal node then we set $p_{k,l} = r/\overline D - \sqrt{r/(\overline D n)}$, if either $k$ or $l$ is the optimal node then $p_{k,l} = r/\overline D$. For the remaining nodes $k$ and $l$, we define $p_{k,l} = 0$. In this situation, $D_\star = \overline D$, $\varepsilon_\star = 0$, and $r_\star = r$ can be chosen arbitrarily. Since in this case the graph structure does not significantly help,\footnote{If we consider only the information we obtain for a node from the graph, and not the information we obtain from \emph{pulling} the node, the error is at least $d\sqrt{\varepsilon/n}$ even at the end of the budget, which is higher than the gap $\sqrt{r\overline D/n}$, and therefore the information coming from the graph structure does not significantly help.} we can use the bound of Theorem~\ref{thls} with the knowledge of which arms are in~$\mathcal S$. The lower bound in this case implies the result when $\sqrt{nr\overline D} \geq an\varepsilon$.  


Alternatively, if $\sqrt{nr\overline D} \leq an\varepsilon$, we consider two cases: $n \leq d$ and $n \geq d$. If $n\leq d$, the result clearly holds because we can, similarly to the proof of Theorem~\ref{thls},  consider an asymmetric graph such that the graph structure does not help and where the detectable dimension is $1$ for $n$ large enough.\footnote{We can set  $p_{l,k_0} =1$ for some suboptimal node $k_0$ and for any $l$; and set $p_{l,k} = r_l/d$ for any $k,l$ and $r_l$ as in the second part of Theorem~\ref{thls}.} If $n \geq d$, we consider a related construction as in the first part of Theorem~\ref{thls}, setting $p_{l,k_0} =1$ for some suboptimal node $k_0$ and any $l$  so that the detectable dimension is $1$ for~$n$ large enough, and setting $p_{l,k} = r_l/d$ for $k\ne k_0$ and any $l$ so that the graph is asymmetric, therefore making the graph structure completely useless. Then, by Theorem~\ref{thls}, the regret is at least of order $\sqrt{drn}$  and since $\sqrt{drn}$ is higher than $\varepsilon n$, we get the result.

\section{Detectable dimension}\label{app:graphs}

In this appendix, we provide additional plots that help understand the behavior
of detectable dimension  $D_\star$ as a function of number of rounds $n$.

\begin{figure}[h]
\begin{center}
\includegraphics[width=0.24\columnwidth]{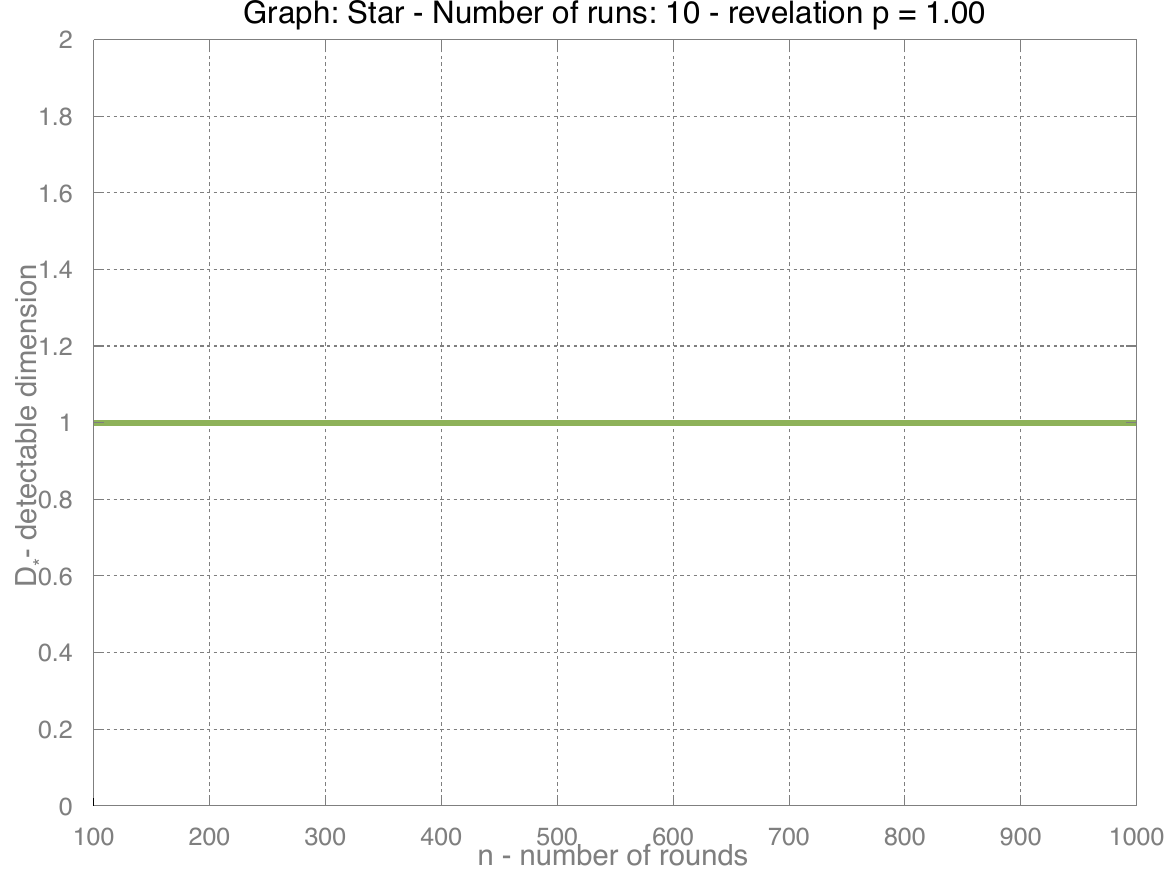}
\includegraphics[width=0.24\columnwidth]{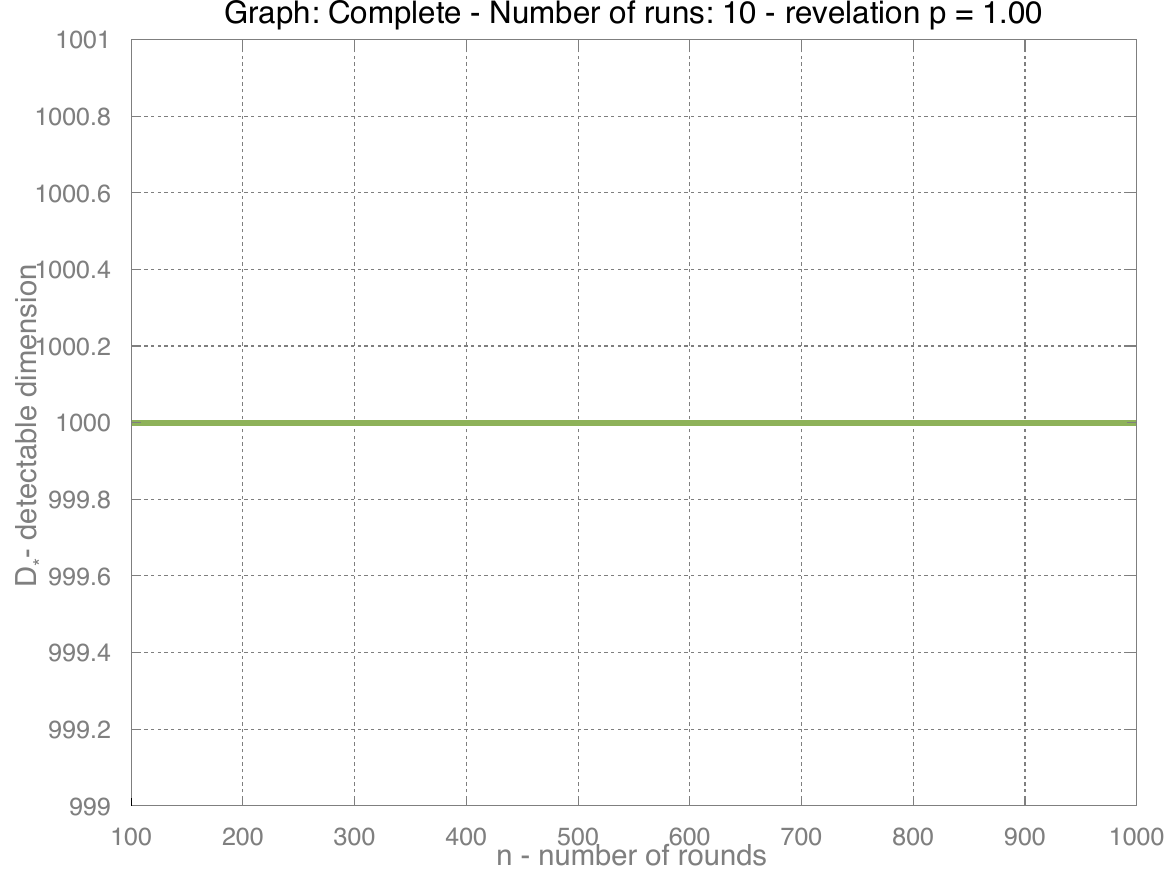}
\caption{%
\textit{Left}: Star graph. 
\textit{Right}: Complete graph.}
\label{fig:ac1}
\end{center}
\end{figure}

As a sanity check, in Figure~\ref{fig:ac1}, we show the behaviour on an easy 
\emph{star} graph where $D_\star =1$  even for a small $n$ and a difficult \emph{complete} graph, where
$D_\star = d$, even for a large $n$. In the case of the empty graph  --- classic bandit setting ---
$D_\star = d$ even for a large $n$ as well.

\begin{figure}[h]
\begin{center}
\includegraphics[width=0.24\columnwidth]{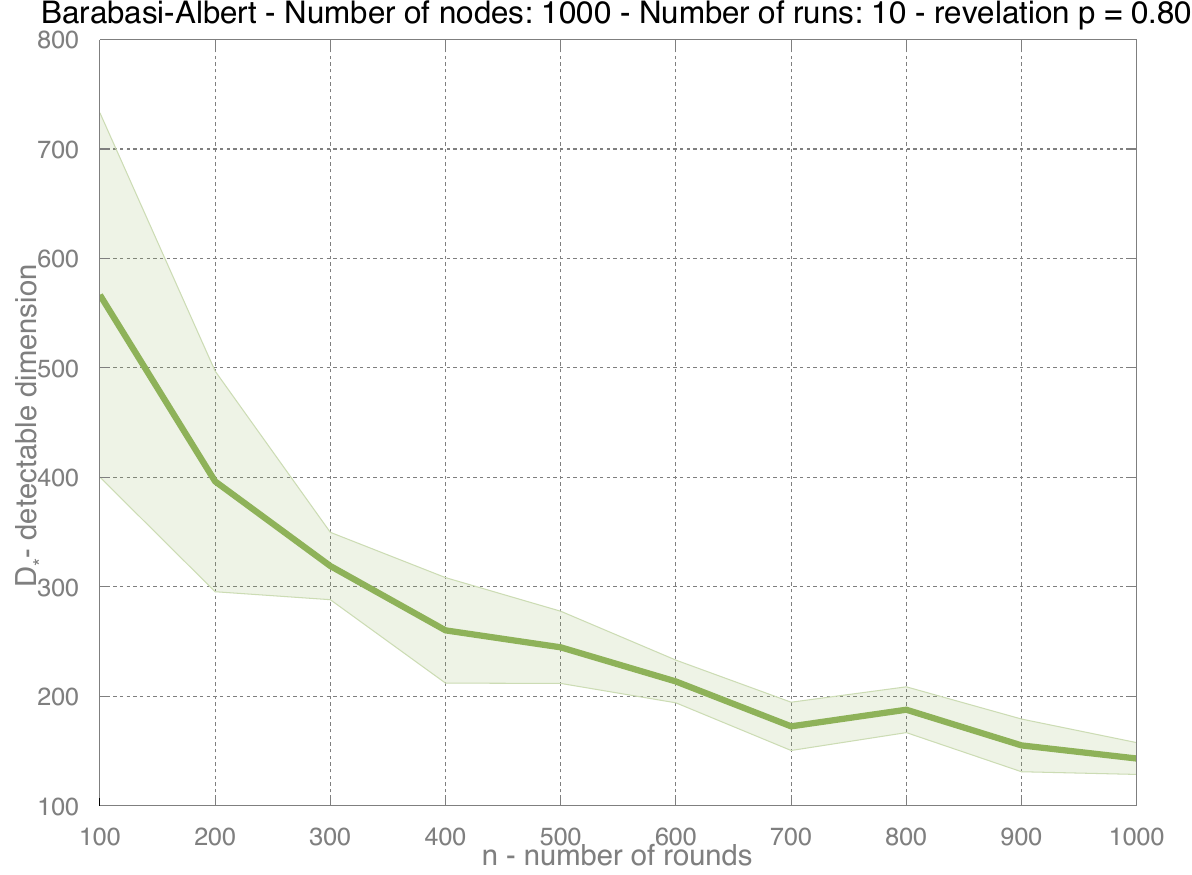}
\includegraphics[width=0.24\columnwidth]{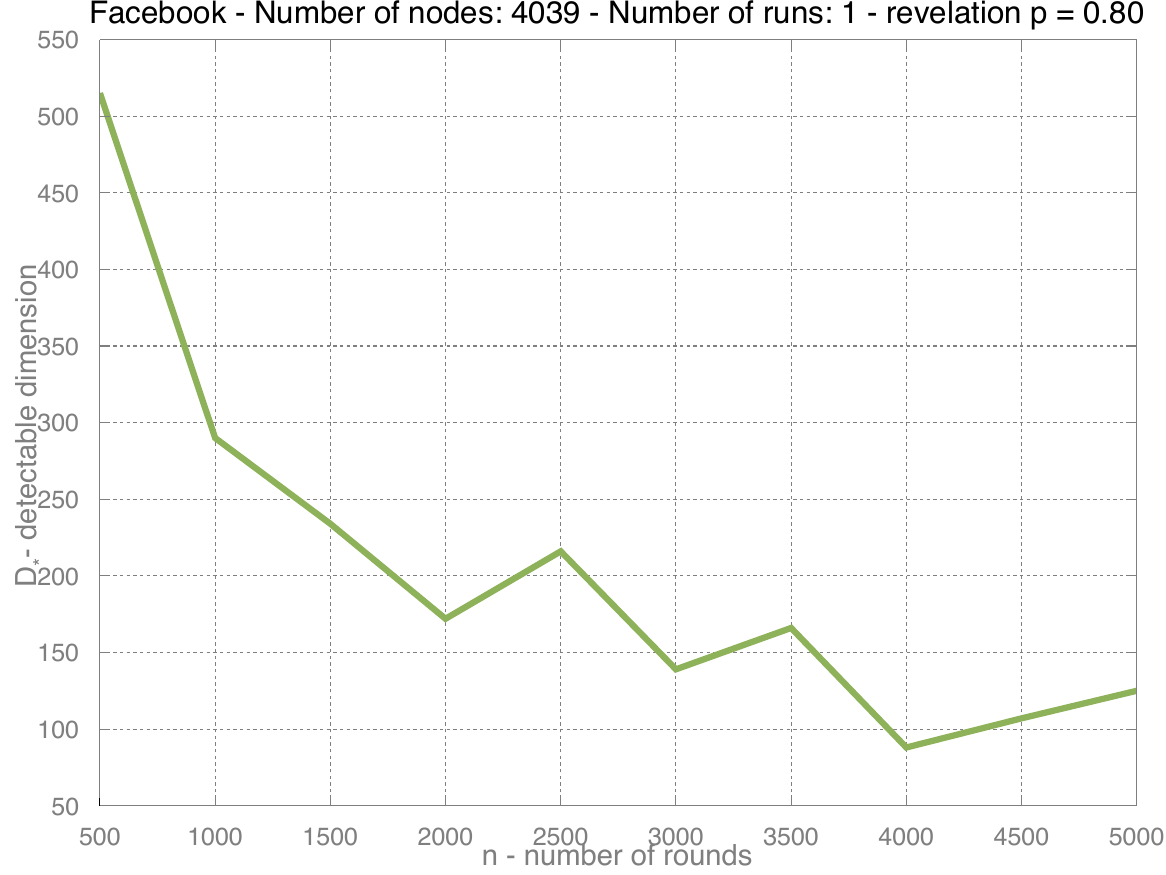}
\includegraphics[width=0.24\columnwidth]{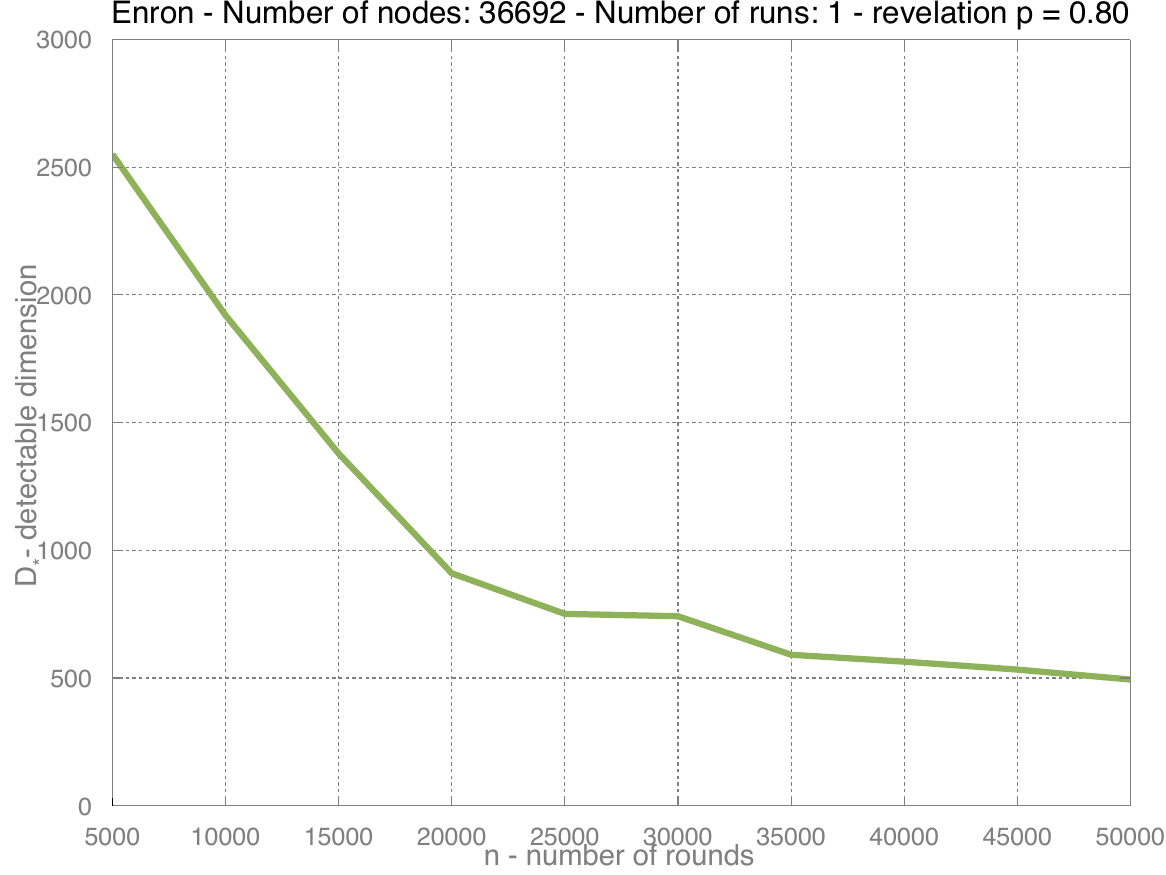}
\includegraphics[width=0.24\columnwidth]{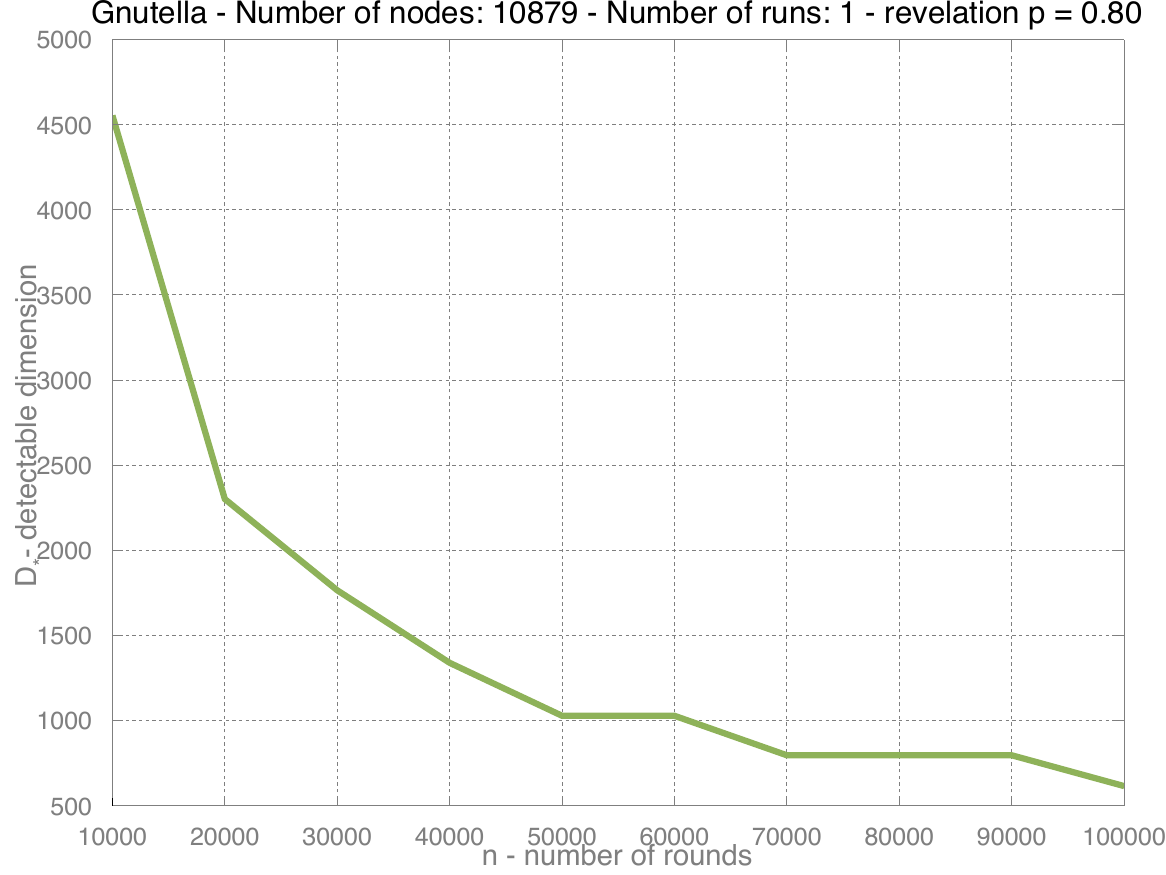}
\caption{%
\textit{Left}: Barab\'asi-Albert. 
\textit{Middle left}: Facebook. 
\textit{Middle right}: Enron. 
\textit{Right}: Gnutella.}
\label{fig:ac2}
\end{center}
\end{figure}

In Figure~\ref{fig:ac2},  we plot  the empirical value of $D_\star$ as a function of $n$, 
for the graphs used in Section~\ref{sec:exp} to give intuition on how $D_\star$ decreases for 
different graphs.

\end{document}